\documentclass[a4paper, 10pt, conference]{ieeeconf}      

\IEEEoverridecommandlockouts                              
\usepackage[a4paper, left=40pt, right=40pt, bottom=68pt, top=68pt]{geometry}
\usepackage[dvipsnames]{xcolor}
\usepackage{amssymb}
\usepackage{amsmath}
\usepackage{commath}

\usepackage{mathtools}
\usepackage{graphicx}

\usepackage{multirow}
\usepackage{adjustbox}
\usepackage{cite}
\usepackage{algorithm}
\usepackage[noend]{algpseudocode}

\title{\LARGE \bf
Magnetic-Assisted Initialization for Infrastructure-free Mobile Robot Localization}

\author{Zhenyu Wu$^{1}$, Mingxing Wen$^{1}$, Guohao Peng$^{1}$, Xiaoyu Tang$^{1}$ and Danwei Wang$^{1}$
\thanks{*This work was supported by the ST Engineering-NTU Corporate Laboratory through the NRF Corporate Laboratory@University Scheme.}
\thanks{$^{1}$All authors are with the School of Electrical and Electronic Engineering, Nanyang Technological University, Singapore 639798. E-mail: {\tt\small zhenyu002@e.ntu.edu.sg}}%
}

\begin{document}

\maketitle
\thispagestyle{empty}
\pagestyle{empty}

\begin{abstract}

Most of the existing mobile robot localization solutions are either heavily dependent on pre-installed infrastructures or having difficulty working in highly repetitive environments which do not have sufficient unique features. To address this problem, we propose a magnetic-assisted initialization approach that enhances the performance of infrastructure-free mobile robot localization in repetitive featureless environments. The proposed system adopts a coarse-to-fine structure, which mainly consists of two parts: magnetic field-based matching and laser scan matching. Firstly, the interpolated magnetic field map is built and the initial pose of the mobile robot is partly determined by the \textit{k}-Nearest Neighbors (\textit{k}-NN) algorithm. Next, with the fusion of prior initial pose information, the robot is localized by laser scan matching more accurately and efficiently. In our experiment, the mobile robot was successfully localized in a featureless rectangular corridor with a success rate of 88$\%$ and an average correct localization time of 6.6 seconds.


\end{abstract}

\section{Introduction}

\begin{figure}[t]
\centering
\includegraphics[width=1.0\linewidth]{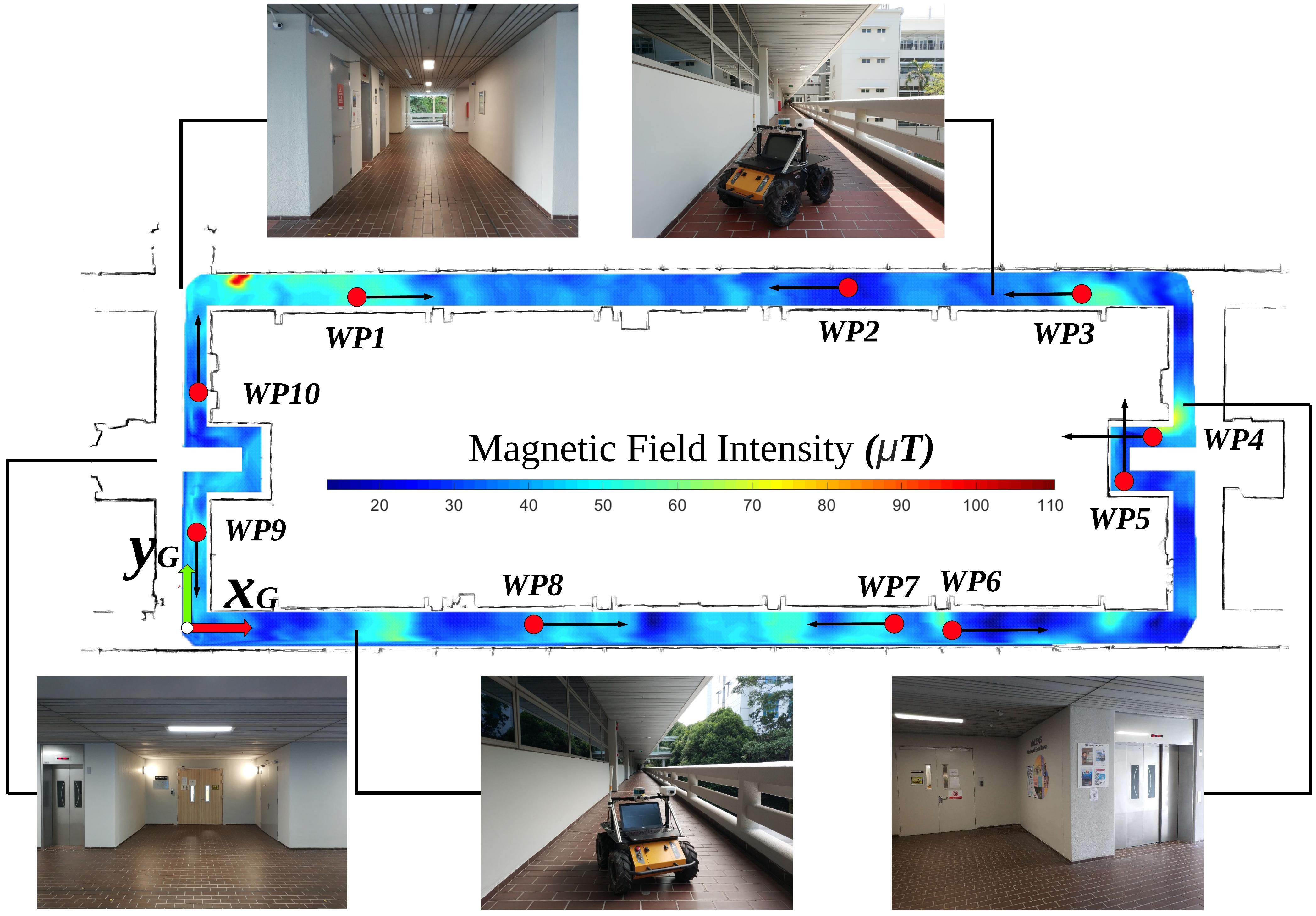}
\caption{The combination of interpolated magnetic field map and occupancy grid map of an approximately 57m$\times$19m featureless rectangular corridor within the S1 building of NTU. Red dots denote ten different starting locations and black arrows represent the heading direction of the robot of the localization process.}
\label{fig:first}
\end{figure}

The development of new algorithms for mobile robot localization is an active area of research in recent years. The existing solutions can be generally classified into two categories: infrastructure-based and infrastructure-free. For infrastructure-based localization, infrastructures like Wi-Fi and RFID are installed to help localize the mobile robot. But usually the costs for purchasing and pre-installing such infrastructures are high. As for infrastructure-free localization, methods such as simultaneous localization and mapping (SLAM)-based and magnetic field-based have much lower cost and can be easily adapted to almost any environment. Normally, SLAM-based approaches work well in feature-rich environments. However, environments like corridors, offices and carparks have lots of repetitive settings with few unique features, making SLAM-based methods inaccurate. Therefore, accurate infrastructure-free localization in such featureless environments is still a challenging issue for mobile robot. An example of such environment is shown in Fig.~\ref{fig:first}.



Magnetic field-based methodology is one of the promising technologies for localization due to its infrastructure-free characteristic and pervasiveness. Ever since \cite{suksakulchai2000mobile} proposed the concept of using magnetic field variations for indoor localization, researchers have demonstrated the feasibility of using the magnetic field alone for both 1-D and 2-D indoor localization \cite{haverinen2009global}, \cite{li2012feasible} and navigation \cite{gozick2011magnetic}, \cite{angermann2012characterization} in a few years. In addition, \cite{haverinen2009global}, \cite{li2012feasible} proved the stability of the disturbed magnetic field over a long period of time. Furthermore, solutions of the SLAM problem were addressed by \cite{vallivaara2011magnetic, jung2015indoor} based on the ambient magnetic field.

The ambient magnetic field consists of the geomagnetic field and the magnetic field generated by ferromagnetic objects. The anomalies caused by ferromagnetic objects, which is generally called \textit{fingerprints}, can be used to describe the environment. The fingerprinting methodology has been becoming dominant in magnetic field localization \cite{li2012feasible}. By applying \textit{k}-NN algorithm, these magnetic fields can be utilized as a medium for mobile robot localization. 

In addition to magnetometer, the Light Detection and Ranging (LiDAR) is often used as a mobile robot localization and mapping sensor. Especially, laser scan matching implemented by LiDAR is generally used\cite{map_fusion, hess2016real, Yue_2019}. Moreover, the scan-to-map matching is adopted due to its high efficiency and robustness rather than the scan-to-scan matching \cite{hess2016real, 8451911}. However, laser scan matching in featureless environments is challenging because it is difficult to collect enough unique environmental features. 


In this paper, we present a magnetic-assisted initialization approach to enhance the performance of infrastructure-free mobile robot localization in repetitive featureless environments. The main contributions of this paper are two-fold:
\begin{itemize}

\item A magnetic-assisted initialization approach, which partly implements \textit{k}-NN based magnetic field matching, has been proposed to improve the performance of infrastructure-free mobile robot localization. 
\item The challenge of infrastructure-free mobile robot localization in repetitive featureless environments has been addressed.

\end{itemize}

The rest of this paper is organized as follows. Related works are reviewed in Section II. Section III introduces framework of the proposed system. Section IV presents experiments, results and evaluations. Finally, conclusions are drawn in Section V.

\section{Related Work}

\subsection{Infrastructure-based \& Infrastructure-free Method}

Most of the existing indoor localization methods are infrastructure-based. Despite its shortcomings like fading due to the high operating frequency, Wireless Area Network (WLAN) technology is still the most common method for indoor localization \cite{li2015using}. However, the access points are deployed for the optimization of communication purposes but not for localization applications \cite{pasku2017magnetic}.

Besides, the infrastructure-based and infrastructure-free method can be integrated. \cite{jung2015indoor} have developed a localization system by utilizing ambient magnetic and radio measurements. By implementing Rao–-Blackwellized particle filter and magnetic-assisted heading correction, the concept of magnetic SLAM was proposed by them. \cite{li2015using} has shown the feasibility of integrating Wi-Fi and magnetic field for indoor localization. However, all of the above mentioned approaches need to pre-install the infrastructures and the localization accuracy is not good enough.

\subsection{Magnetic Field-based Localization}

As the magnetic field is temporally stable and requires no hardware installation, it often yields comparable or even better results to Wi-Fi based localization \cite{le20123}. The approach implemented by \cite{gozick2011magnetic} identifies pillars in a floor as landmarks thus creating magnetic field maps for the whole floor. \cite{angermann2012characterization} have shown that the change in the tri-axial magnetic field vector is sufficient to represent re-recognizable features based upon which precise localization can be performed.

In addition, many works have shown the applicability of magnetic field to localization issues for both mobile robots \cite{haverinen2009global}, \cite{angermann2012characterization}, \cite{vallivaara2011magnetic}, \cite{frassl2013magnetic} and pedestrians \cite{haverinen2009global}, \cite{gozick2011magnetic}, \cite{le20123, frassl2013magnetic}. In \cite{frassl2013magnetic}, M. Frassl \textit{et al.} further expanded their work in \cite{angermann2012characterization} to show how the use of the magnetic field demonstrates significant improvements over odometry-based approach with respect to the localization accuracy for both robots and pedestrians.

\section{System Overview}

\begin{figure}[t]
\centering
\includegraphics[scale=0.5]{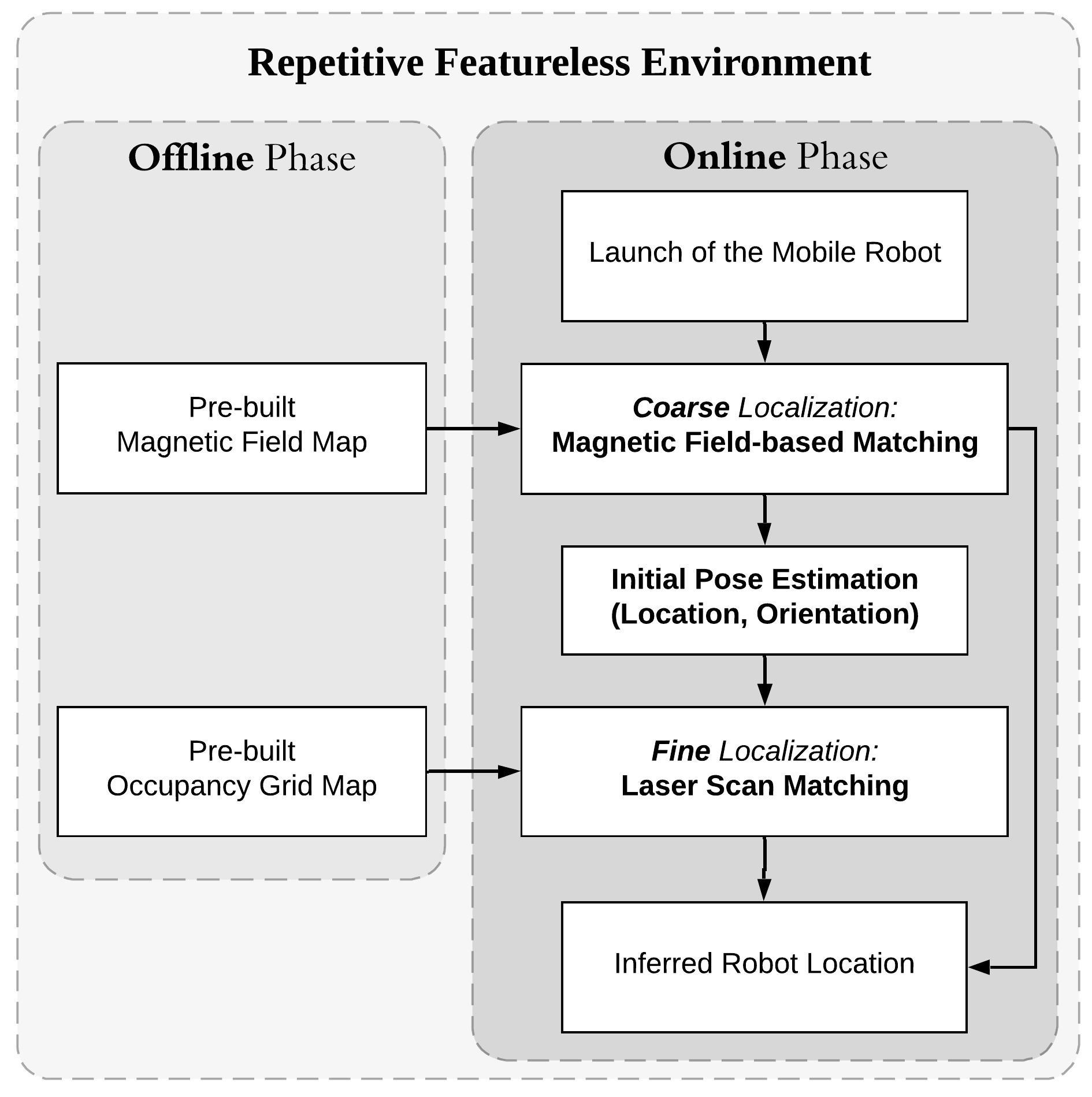}
\caption{The proposed system flowchart}
\label{fig:system}
\end{figure}

As shown in the flowchart in Fig.~\ref{fig:system}, the proposed system is mainly composed of three parts: 1) Coarse Localization: Magnetic Field-based Matching; 2) Initial Pose Estimation; 3) Fine Localization: Laser Scan Matching. Details of these three parts are discussed below.

\subsection{System Framework}
The whole system is mainly divided into two phases: the \textit{Offline Phase} and the \textit{Online Phase}. For the \textit{Offline Phase}, its main objective is to collect magnetic field fingerprints, build interpolated magnetic field map and occupancy grid map of the environment. For the \textit{Online Phase}, its main function is to fuse the magnetic field-based matching and laser scan matching to infer the accurate robot location. 

As can be seen in the flowchart, what connects the magnetic field-based matching and the laser scan matching is the initial pose estimation. The initial pose information consists of two parts: the Location ${(x, y)}$ and Orientation ${\theta}$. Both the location and orientation information are provided by the results of the magnetic field-based matching. Then initial pose information is fused with the laser scan matching to enhance the performance of infrastructure-free mobile robot localization in repetitive featureless environments.

\subsection{Coarse Localization: Magnetic Field-based Matching}
Normally, the magnetic field-based matching algorithm is divided into two phases: \textit{Offline Phase} and \textit{Online Phase}. The main objective of the \textit{Offline Phase} is to build the interpolated magnetic field map. This map associate a 2-D location with a 3-D magnetic field vector $B$. In the \textit{Online Phase}, the real-time magnetometer measurements are obtained and the coarse location of the robot can be inferred by applying the \textit{k}-NN algorithm.

\subsubsection{Magnetic Field Fingerprints Collection}
In general, the magnetic field vector consists of three components: $B_x$, $B_y$, and $B_z$, which represent intensities in $X$, $Y$ and $Z$ directions respectively. In each fingerprint, the localization accuracy is higher if more components are used. Thus, using all three components is preferred over using the magnitude alone. The magnetic field fingerprint database is built with the mobile robot teleoperated on different desired routes, which are along the trajectories parallel to the straight fence or wall of the corridor. By recording both 3-D magnetic field vectors and 2-D location information, the magnetic field fingerprint database can be constructed.

\subsubsection{Bilinear Interpolation of the Magnetic Field}
\begin{figure}[t]
\centering
\includegraphics[width=1.0\linewidth]{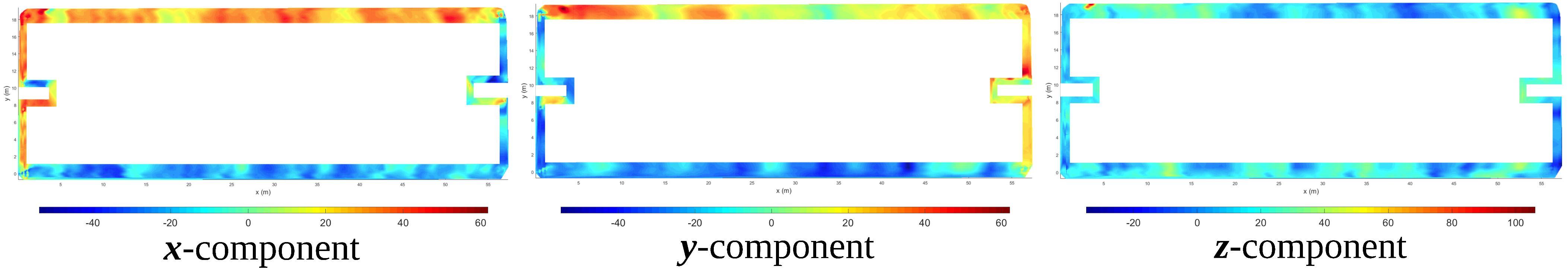}
\caption{Magnetic field vector components of the rectangular corridor}
\label{fig:mag_xyz}
\end{figure}

The cost of collecting complete magnetic field vectors is high due to the narrow measurement range of the magnetometer and the large size of the environment. It is impossible for the robot to go to every place in the environment so the magnetic field fingerprint measurements can be sparse or even vacant in some places. In order to solve this problem, the bilinear interpolation method is implemented to estimate the missing magnetic field intensities at unexplored locations and build the magnetic field map of the environment. The interpolated magnetic field vector components of the counter-clockwise robot routes are shown in Fig.~\ref{fig:mag_xyz}. 

\begin{figure}[t]
\centering
\includegraphics[scale=0.35]{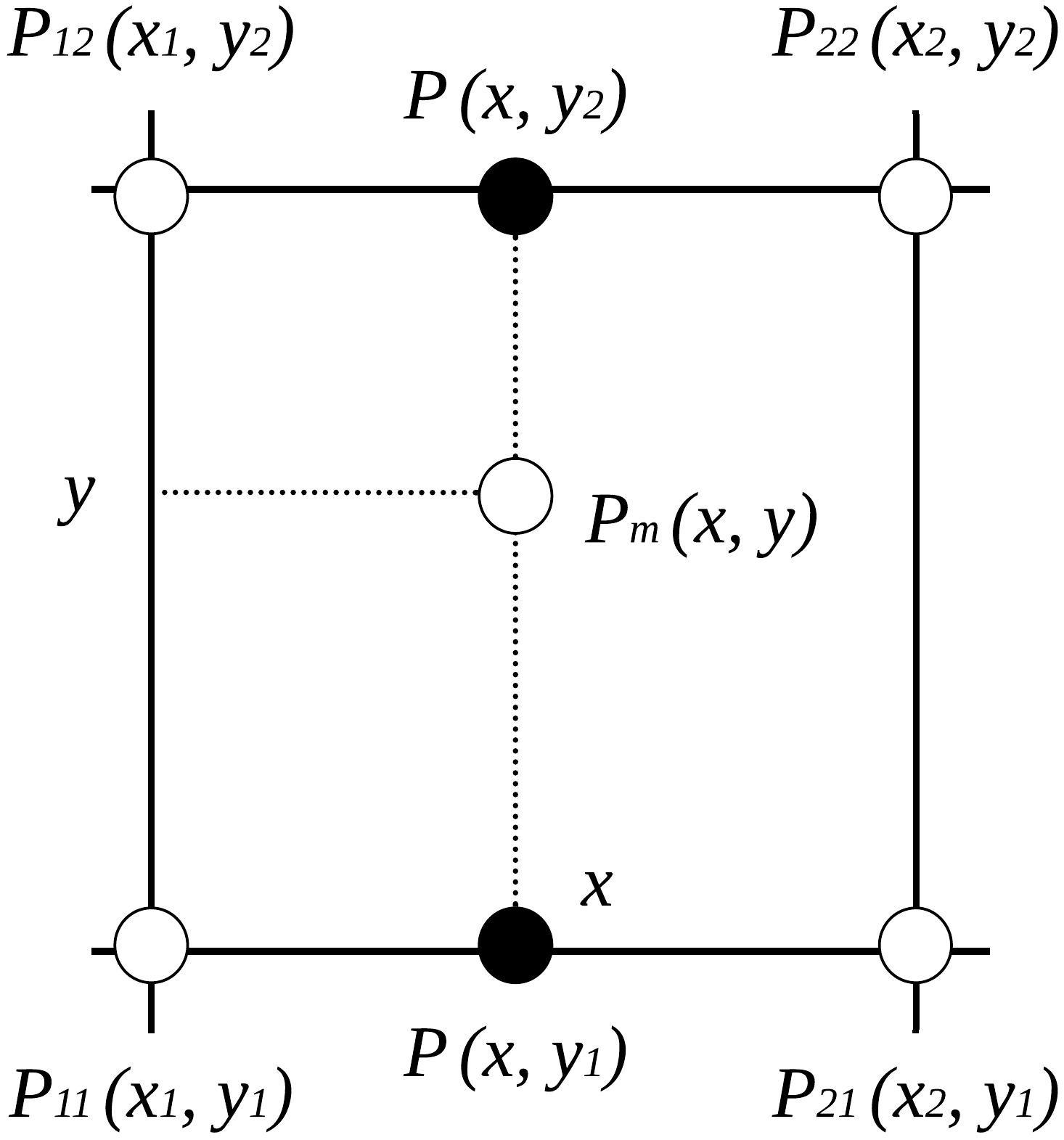}
\caption{Bilinear interpolation of the grid map. Point $P_m$ is the point whose magnetic field intensity shall be interpolated.}
\label{fig:bilinear}
\end{figure}

To approximate the magnetic field intensity vector $B(P_m)$ of an observation point, the coordinates of the observation point $P_m$ and the coordinates of the four closest reference points $(P_{11}, P_{12}, P_{21}, P_{22})$ are utilized. As illustrated in Fig.~\ref{fig:bilinear}, the linear interpolation along x-axis yields:

\begin{equation} \label{eq:1}
B(P(x,y_1)){\,}{\approx}{\,}{\frac{{x_2}-x}{{x_2}-{x_1}}}B(P_{11})+{\frac{x-{x_1}}{{x_2}-{x_1}}}B(P_{21})
\end{equation}
\begin{equation} \label{eq:2}
B(P(x,y_2)){\,}{\approx}{\,}{\frac{{x_2}-x}{{x_2}-{x_1}}}B(P_{12})+{\frac{x-{x_1}}{{x_2}-{x_1}}}B(P_{22})
\end{equation}

Then the linear interpolation along y-axis can be yielded as \cite{kohlbrecher2011flexible}:
\begin{align} \label{eq:3}
\begin{split}
B(P_m){\,}{\approx}{}&{\,}{\frac{{y_2}-y}{{y_2}-{y_1}}}B(P(x,y_1))+{\frac{y-{y_1}}{{y_2}-{y_1}}}B(P(x,y_2)) \\
={}& {\frac{{y_2}-y}{{y_2}-{y_1}}}\bigg({\frac{{x_2}-x}{{x_2}-{x_1}}}B(P_{11})+{\frac{x-{x_1}}{{x_2}-{x_1}}}B(P_{21}) \bigg) \\
+{}& {\frac{y-{y_1}}{{y_2}-{y_1}}}\bigg({\frac{{x_2}-x}{{x_2}-{x_1}}}B(P_{12})+{\frac{x-{x_1}}{{x_2}-{x_1}}}B(P_{22}) \bigg)
\end{split}
\end{align}

\subsubsection{\textit{k}-NN Based Matching}

After the previous section \textit{1)} and \textit{2)}, the interpolated magnetic field map is obtained. Thus the \textit{k}-NN algorithm is implemented to accomplish the coarse robot location inference during the online phase. The \textit{k}-NN algorithm uses two datasets, which are the training set (magnetic field fingerprint database) and the testing set (real-time magnetometer measurements). Firstly, the members in training set are described by 3-D magnetic field vector and corresponding 2-D location. A point in 3-D vector space, which is established by the magnetic field vector, is represented by each member in training set. Thus in such a way, all of the members are mapped into a 3-D vector space. Secondly, the members in testing set are also distributed in the 3-D vector space but without knowing their location. After a testing sample is given, the \textit{k}-NN classifier searches the \textit{k} members in training set which are closest to the given sample. Then the location which has the majority number in the \textit{k} neighbors is assigned to the testing sample. The closeness between the testing sample and the member in training set can be defined in terms of the \textit{Euclidean Distance}, which is given as follows:

\begin{equation} \label{eq:6}
d \big( m_i, m_j \big) = \bigg( \sum\limits_{l=1}^{n} |{m_{il} - m_{jl}}|^2 \bigg)^{\frac{1}{2}}
\end{equation}
where $m_i$ and $m_j$ represent two members in n-dimensional vector space; $n$ is the dimension of the vector space; ${m_{il}}$ and ${m_{jl}}$ indicate the intensities of the magnetic field for the two members in $X/Y/Z$ direction respectively. 


\subsubsection{Orientation Determination}
\begin{figure}[t]
\centering
\includegraphics[scale=0.4]{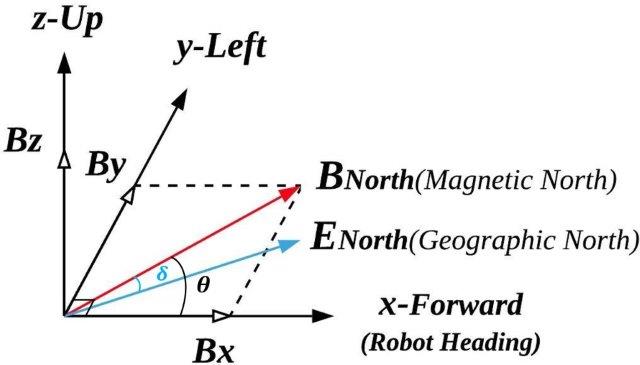}
\caption{Magnetic field directions}
\label{fig:direction}
\end{figure}

In our case, the $B_x$ represents the magnetic field strength in the heading direction of the robot. $B_y$ represents the magnetic field strength in counter clockwise 90 degrees from the $B_x$ direction. $\delta$ represents the magnetic declination angle between the magnetic north pole and geographic north pole. And $\theta$ represents the angle between the heading of the robot and the magnetic north. Suppose the angle between the heading of the robot and the geographic north is $\theta_d$, then it can be determined by ${{\theta_d}= {\theta}-{\delta}}$. The magnetic declination angle $\delta$ in Singapore roughly equals to zero so the magnetic north pole is in the same direction with the geographic north pole. This means the vector sum of the $B_x$ and $B_y$ is pointing to the \textit{Geographic North}. The above mentioned relationships between these directions are shown in Fig.~\ref{fig:direction}.

Therefore, the rotation from the robot body frame to East-North-Up global reference frame (\textit{Right-handed Cartesian Coordinate System}) is taken as the orientation estimation for the robot. By measuring ambient magnetic field vector in these two frames, the rotation between these two frames can be predicted. Leaving out the translation part, the origins of these two frames are supposed to coincide with each other. The ambient magnetic field vector $B \in \mathbb{R}^3$ is written as $B_G$ in the static global reference frame and accordingly the $B$ in the robot body frame can be represented as $B_L$. With respect to the global reference frame, the heading of the robot body frame is expressed as a rotation matrix ${\prescript{L}{G}{R}}$ which can be represented as ${B_G}={\prescript{L}{G}{R}}\cdot{B_L}$. The angle ${\phi}$ and unified axis $\mathbf{a}$ of the rotation are derived as:

\begin{align}\label{eq:7}
{\phi} = {}& {\arccos{\Bigg(\frac{{B_L}{\,}{\cdot}{\,}{B_G}}{{\norm{B_L}}{\,}{\norm{B_G}}}\Bigg)}}\\
\label{eq:8}
\mathbf{a} = {}& \frac{{B_L}{\,}{\times}{\,}{B_G}}{\norm{{B_L}{\,}{\times}{\,}{B_G}}}
\end{align}

In our case, it is reasonable to assume that the robot is moving along the direction parallel to the straight fence or wall of the corridor as most of the warehouse robots are doing so in corridors. The $x$ and $y$ axis of the global reference frame are shown as $x_G$ and $y_G$ in Fig.~\ref{fig:first}. Thus the determination of the robot orientation can be divided into three cases: 1) Angle ${\phi=0}$, the robot is heading towards positive $x$ axis; 2) Angle ${\phi=\pi}$, the robot is heading towards negative $x$ axis; 3) Angle ${\phi={\frac{\pi}{2}}}$, the robot is heading towards negative $y$ axis if unified axis $\mathbf{a}$ pointing upwards or the robot is heading towards positive $y$ axis if unified axis $\mathbf{a}$ pointing downwards.

\subsection{Fine Localization: Laser Scan Matching}
For the fine localization part, the laser scan matching method is mainly composed of three parts: Local 2-D SLAM, Global Loop Closure, and Localization. For the first two parts, both of them optimize the scan pose ${\sigma = \big(\sigma_{x}, \sigma_{y}, {\psi})^T}$ which consist of a translation ${\big(\sigma_{x}, \sigma_{y})^T}$ and a rotation ${\psi}$ of the LiDAR scans.

\subsubsection{Local 2-D SLAM}

Submap, which is a small part of the world, is built by the iterative process of repeatedly aligning a few consecutive laser scans with submap coordinate frames. By setting the origin of the laser scan at $(0,0)^T$ in the \textit{Euclidean Plane} $\mathbb{R}^2$, the laser scan endpoints can be represented as ${S = {\{(s_{j,x},s_{j,y})^T}\}_{j=1,2,...,J}}$, $s_j = (s_{j,x},s_{j,y})^T \in \mathbb{R}^2$. The real-time laser scan endpoints are transformed from the scan frame into the submap frame by the transformation $T_{\sigma}$. This process is defined as:

\begin{equation} \label{eq:9}
T_{\sigma}{s_j} = 
\underbrace{
\left( \begin{array}{cc}
cos{(\psi)} & -sin{(\psi)} \\
sin{(\psi)} & cos{(\psi)} \end{array} \right)}_{R_{\sigma}}{\left( \begin{array}{cc}s_{j,x} \\
s_{j,y} \end{array} \right)} + \underbrace{
\left( \begin{array}{cc}\sigma_{x} \\
\sigma_{y} \end{array} \right)}_{t_{\sigma}}
\end{equation}
where $R_{\sigma}$ and $t_{\sigma}$ represent the Rotation and Translation part respectively. 


According to the current local submap which implements the scan matcher \cite{agarwal2012ceres}, the scan pose $\sigma$ is optimized before the insertion of a laser scan into a submap. Function of the scan matcher is to search for a optimum scan pose which can best align the current laser scan with the submap once every few seconds. This can be represented as a nonlinear least squares problem: 

\begin{equation} \label{eq:10}
{\sigma}^* = \operatorname*{argmin}_\sigma  \sum_{j=1}^{J}\Big( {1-M_{smooth}}\big({T_{\sigma}}{s_j}\big)\Big)^{2}
\end{equation}
where function $M_{smooth}$ : $\mathbb{R}^2$ $\rightarrow$ $\mathbb{R}$ smooths the probability values at the coordinates given by $\big({T_{\sigma}}{s_j}\big)$ in the local submap; $T_{\sigma}$ is the previously mentioned transformation. 


\subsubsection{Global Loop Closure}
Since larger spaces are composed of many small submaps, the sparse pose adjustment \cite{konolige2010efficient} is adopted for the optimization of all laser scan poses and submap poses. Once the submap is settled, the pairs made up of a submap and a scan is considered for loop closure. The scan matcher mentioned in the previous subsection, which is performing in the background, will add the corresponding pose to the optimization problem once it finds a good match. Like the scan matching, a nonlinear least squares model also can be applied to solve the global loop closure optimization problem. The extensive literature on this part is available in \cite{konolige2010efficient},\cite{clausen1999branch}. For the global approach, the loop closure removes the accumulated laser scan matching error caused by the local approach.

\subsubsection{Localization}
As mentioned in the previous two subsections, the occupancy grid map is built by continuously accumulating submaps and implementing loop closure optimizations. With the initial pose information provided by the magnetic field-based matching, the mobile robot can be localized by laser scan matching more accurately. Once the real-time laser scans are correctly matched with the points and lines in the previously built occupancy grid map, the mobile robot is considered to be successfully localized. 

Repetitive environments like corridors and offices have lots of highly similar features. Localization in such environments often leads to failed matching due to the lack of unique features. To make this issue more clear, an example of the laser scan matching in such environment is shown in Fig.~\ref{fig:comparison}. Without any initial information, a failed matching is shown in Fig.~\ref{fig:comparison}(b) where the robot was wrongly localized to a very similar turning corner of the corridor. But if both the initial location ${(x, y)}$ and orientation ${\theta}$ information provided by the magnetic field-based matching are given to the laser scan matching, the matching will be successful in most cases. A successful matching example is shown in Fig.~\ref{fig:comparison}(c).




\section{Experimental Results}

\subsection{Platform}
\begin{figure}[b]
\centering
\includegraphics[scale=0.3]{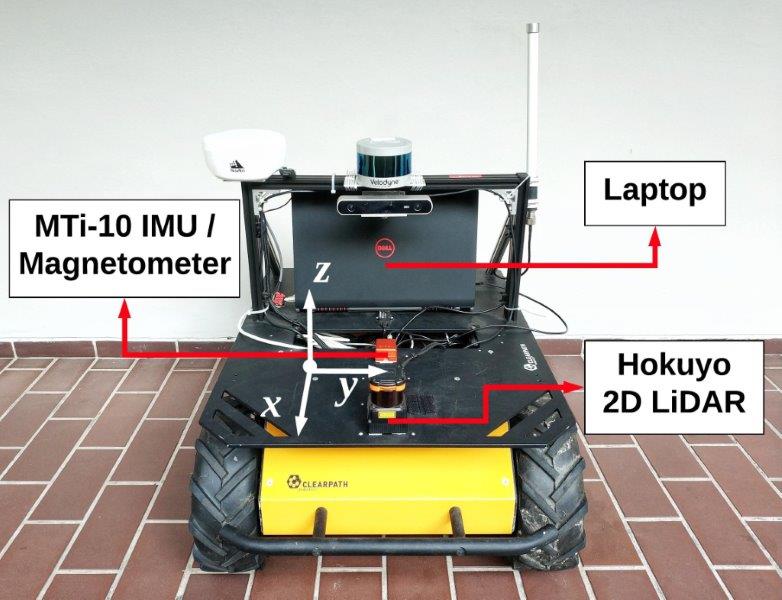}
\caption{Huksy robot platform}
\label{fig:platform}
\end{figure}

Fig.~\ref{fig:platform} demonstrates the experimental platform used in this study and the robot body frame. The main part of the platform is a Husky A200 Robot. A Xsens MTi-10 IMU is installed at the central place of the robot. And a Hokuyo UTM-30LX 2-D Scanning Laser Rangefinder is installed at the front of the robot. All the tests are performed on a laptop with AMD FX-9830P CPU @3.0GHz  and a RAM of 12GB size.

\subsection{Magnetic Field Map \& Occupancy Grid Map Building}
The experiments are conducted in a featureless rectangular corridor environment with offices in the central part. The method \cite{hess2016real}, which is called Google \textit{Cartographer}, is implemented to build the 2-D occupancy grid map. The interpolated magnetic field map combined with the occupancy grid map of the corridor environment is shown in Fig.~\ref{fig:first}. Three different desired robot routes are conducted to cover as much area of the corridor as possible. Each route goes around the corridor environment in one clockwise circle and in another one counter-clockwise circle. The width of the corridor in the environment is around 2m and the interval between each route is 38 cm. 

\subsection{Localization}
Three different initial pose options are compared in the localization part, which are: 1) Initial pose with location and orientation; 2) Initial pose with only location; 3) Without initial pose. For the second option, the orientation data is set to default zero. For each initial pose option of each waypoint, ten individual random trials are conducted.

\begin{figure}[t]
\centering
\includegraphics[width=1.0\linewidth]{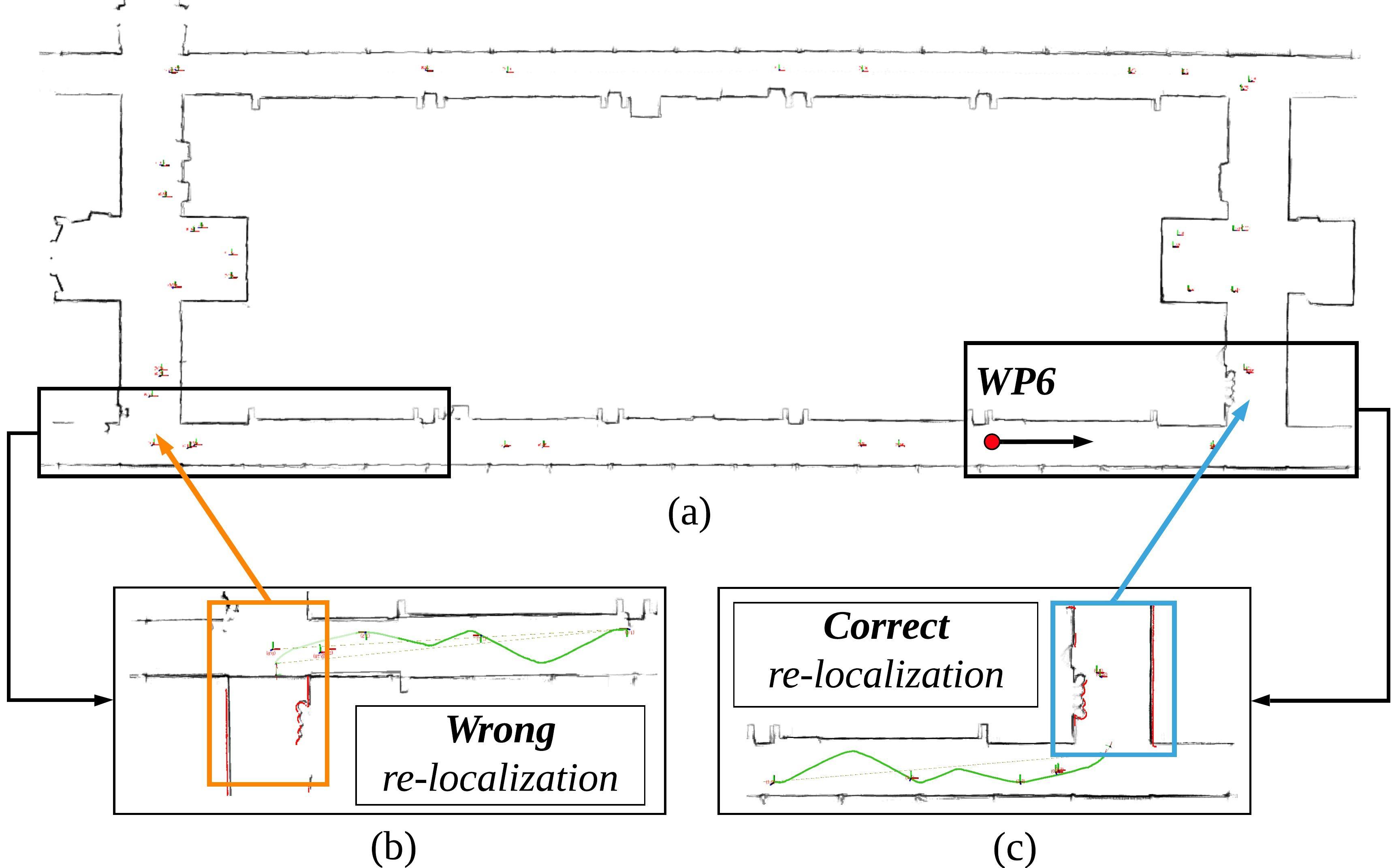}
\caption{The laser scan matching result of waypoint {\#}6  for two different initial pose options. The green curve and red line denote the robot trajectory and the real-time laser scan respectively, and the small RGB coordinates represent the submap coordinate frames: (a) The occupancy grid map of the corridor environment; (b) Localization without initial pose---Failed; (c) Localization with initial location and orientation---Successful.}
\label{fig:comparison}
\end{figure}

\begin{figure}[b]
\centering
\includegraphics[width=1.0\linewidth]{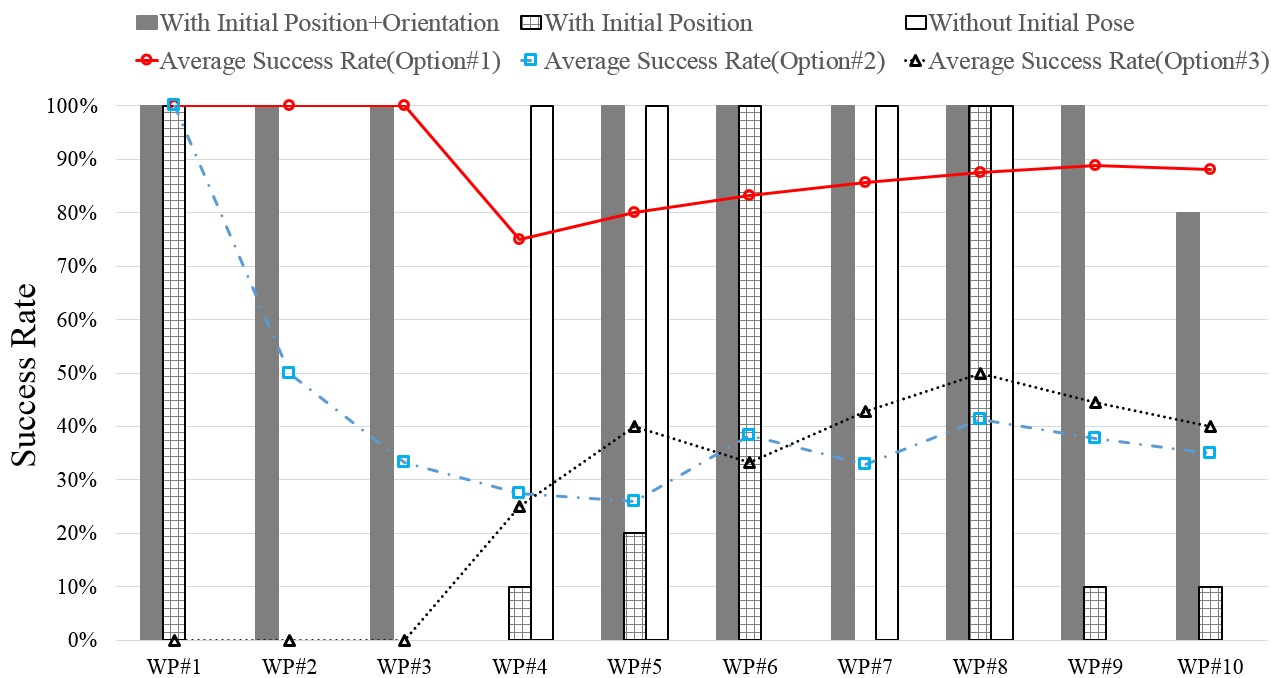}
\caption{Localization success rate for different initial pose options}
\label{fig:success}
\end{figure}

The localization results of waypoint {\#}6 is shown in Fig.~\ref{fig:comparison} as an illustration of the successful and failed scenarios, where "waypoint" is abbreviated as "WP". And the localization success rate in this corridor environment is shown in Fig.~\ref{fig:success}. The histograms represent the success rate for each initial pose option at each waypoint. The lines represent the cumulative success rate of different initial pose options. The proposed method has the highest average success rate of \textbf{88$\boldsymbol{\%}$} over all waypoints, while the option with only location information and the option without any initial information achieve an average success rate of 35$\%$ and 40$\%$ respectively over all waypoints. The sudden drop of the success rate of waypoint {\#}4 is due to the strong magnetic interference from a nearby cargo elevator (as shown in Fig.~\ref{fig:first}). Wrong orientation information is given to the robot thus it worsens the localization performance. 

As shown in Table I, the localization efficiency is measured by the time the robot takes to be successfully localized and the accuracy is measured by the \textit{Root-Mean-Square Error(RMSE)}. The $F$ in Table I represents that the robot can not localize itself so the experiment is considered failed. The proposed method achieves an average correct localization time of \textbf{6.6 seconds}, while the option with only location information and the option without any initial information achieve an average of 7.17 seconds and 11.03 seconds respectively. As for the accuracy, the RMSE of the proposed method is \textbf{1.43m} over all successfully localized waypoints, while the RMSE of the option with only known initial location and the option with unknown initial pose are 2.26m and 2.25m respectively.

The results show clearly that the proposed method has greatly improved the localization success rate and efficiency compared with only using the laser scan matching-based localization. And the results also reveal that even without initial pose information, the localization performance is comparable with the performance of the initial pose option with only location. This demonstrates that orientation plays a more important role than location in the initialization of infrastructure-free mobile robot localization.

\begin{table}[h]
\centering
\caption{Localization Efficiency and Accuracy of Three Initial Pose Options}
\label{tab:my-table}
\begin{adjustbox}{width=0.42\textwidth}
\small
\begin{tabular}{cccccc}
\hline
\multirow{2}{*}{Waypoint} & \multirow{2}{*}{Initial Pose Option} & \multicolumn{3}{c}{Correct Localization Time} & \multirow{2}{*}{RMSE} \\ \cline{3-5}
                          &                                      & Fastest       & Slowest       & Average          &                       \\ \hline
\multirow{3}{*}{\#1}      & 1                                    & 5.6s          & 7.1s          & \textbf{6.1s}    & \textbf{0.14m}        \\
                          & 2                                    & 5.8s          & 6.5s          & 6.2s             & 0.17m                 \\
                          & 3                                    & \textit{F}    & \textit{F}    & \textit{F}       & \textit{F}            \\ \hline
\multirow{3}{*}{\#2}      & 1                                    & 4.9s          & 7.9s          & \textbf{5.7s}    & \textbf{0.27m}        \\
                          & 2                                    & \textit{F}    & \textit{F}    & \textit{F}       & \textit{F}            \\
                          & 3                                    & \textit{F}    & \textit{F}    & \textit{F}       & \textit{F}            \\ \hline
\multirow{3}{*}{\#3}      & 1                                    & 4.9s          & 7.9s          & \textbf{7.1s}    & \textbf{0.5m}         \\
                          & 2                                    & \textit{F}    & \textit{F}    & \textit{F}       & \textit{F}            \\
                          & 3                                    & \textit{F}    & \textit{F}    & \textit{F}       & \textit{F}            \\ \hline
\multirow{3}{*}{\#4}      & 1                                    & \textit{F}    & \textit{F}    & \textit{F}       & \textit{F}            \\
                          & 2                                    & \textit{F}    & \textit{F}    & 38s              & 1.05m                 \\
                          & 3                                    & 6.6s          & 7.5s          & \textbf{6.9s}    & \textbf{0.72m}        \\ \hline
\multirow{3}{*}{\#5}      & 1                                    & 4.7s          & 6.7s          & \textbf{5.5s}    & \textbf{0.42m}        \\
                          & 2                                    & 25s           & 34s           & 29.5s            & 0.63m                 \\
                          & 3                                    & 8.4s          & 12s           & 9.2s             & 0.71m                 \\ \hline
\multirow{3}{*}{\#6}      & 1                                    & 4.9s          & 7.9s          & \textbf{5.8s}    & \textbf{0.10m}        \\
                          & 2                                    & 5.2s          & 6.6s          & 5.9s             & 0.45m                 \\
                          & 3                                    & \textit{F}    & \textit{F}    & \textit{F}       & \textit{F}            \\ \hline
\multirow{3}{*}{\#7}      & 1                                    & 7.8s          & 9.5s          & \textbf{8.6s}    & \textbf{0.50m}        \\
                          & 2                                    & \textit{F}    & \textit{F}    & \textit{F}       & \textit{F}            \\
                          & 3                                    & 10s           & 14.5s         & 11.2s            & 0.65m                 \\ \hline
\multirow{3}{*}{\#8}      & 1                                    & 7.6s          & 11.4s         & \textbf{9.2s}    & \textbf{0.02m}        \\
                          & 2                                    & 7.3s          & 11.5s         & 9.4s             & 0.3m                  \\
                          & 3                                    & 15.1s         & 19.5s         & 16.8s            & 0.80m                 \\ \hline
\multirow{3}{*}{\#9}      & 1                                    & 5.3s          & 5.7s          & \textbf{5.4s}    & \textbf{0.41m}        \\
                          & 2                                    & \textit{F}    & \textit{F}    & 25s              & 0.54m                 \\
                          & 3                                    & \textit{F}    & \textit{F}    & \textit{F}       & \textit{F}            \\ \hline
\multirow{3}{*}{\#10}     & 1                                    & 5.3s          & 6.2s          & \textbf{5.7s}    & \textbf{0.95m}        \\
                          & 2                                    & \textit{F}    & \textit{F}    & 36s              & 1.03m                 \\
                          & 3                                    & \textit{F}    & \textit{F}    & \textit{F}       & \textit{F}            \\ \hline

\end{tabular}
\end{adjustbox}
\end{table}


\section{CONCLUSIONS}

This paper presents a magnetic-assisted initialization approach to enhance the performance of infrastructure-free mobile robot localization in repetitive featureless environments. The fusion of magnetic field-based matching and laser scan matching is realized in the proposed system. The initial pose of the mobile robot is partly determined by the \textit{k}-NN algorithm. Then with the fusion of prior initial pose information, the mobile robot is localized by laser scan matching more accurately and efficiently. The experimental results indicate that the proposed initial pose option has the highest localization success rate of 88$\%$ while the option with only location information and the option without any initial information achieve an average success rate of 35$\%$ and 40$\%$ respectively. And the RMSE of the proposed method is 1.43 m. The proposed approach has demonstrated a significant improvement on robot localization robustness and efficiency, especially for repetitive featureless environments. Future development work will focus on developing better magnetic field modeling methods in order to provide more accurate initial information for mobile robot localization.





\addtolength{\textheight}{-12cm}   

\bibliographystyle{IEEEtran}
\bibliography{Citations}

\end{document}